\RequirePackage{fix-cm}
\documentclass[smallextended]{svjour3}       
\smartqed  
\usepackage{graphicx}
\usepackage{float}
\usepackage{subfigure}
\usepackage{mathtools}
\usepackage{amssymb}
\usepackage{geometry}
\begin{document}

\title{Proximal Policy Optimization Smoothed Algorithm
}


\author{Wangshu Zhu         \and
        Andre Rosendo 
}


\institute{Wangshu Zhu \at
              \email{zhuwsh@shanghaitech.edu.cn}           
           \and
           Andre Rosendo \at
              \email{arosendo@shanghaitech.edu.cn} 
}

\date{Received: date / Accepted: date}

\maketitle

\begin{abstract}
Proximal policy optimization (PPO) has yielded state-of-the-art results in policy search, a subfield of reinforcement learning, with one of its key points being the use of a surrogate objective function to restrict the step size at each policy update. Although such restriction is helpful, the algorithm still suffers from performance instability and optimization inefficiency from the sudden flattening of the curve. 
To address this issue we present a PPO variant, named Proximal Policy Optimization Smooth Algorithm (PPOS), and its critical improvement is the use of a functional clipping method instead of a flat clipping method. We compare our method with PPO and PPORB, which adopts a rollback clipping method, and prove that our method can conduct more accurate updates at each time step than other PPO methods. Moreover, we show that it outperforms the latest PPO variants on both performance and stability in challenging continuous control tasks.
\end{abstract}

\section{Introduction}
\label{intro}
Reinforcement learning, especially deep model-free reinforcement learning, has achieved great progress in recent years. The application of reinforcement learning ranges from video games \cite{Ref2} \cite{Ref1} , board games \cite{Ref3}, robotics \cite{Ref4} \cite{Ref5}, and challenging control tasks \cite{Ref7} \cite{Ref6}. Within these approaches, Policy gradient (PG) methods are widely used in model-free policy search algorithms, contributing significantly to the fields of robotics and high-dimensional control. PG methods update the policy with a step estimator of the gradient of the expected return and gradually converge to the optimal policy \cite{Ref8}. One of the most significant problems of PG-based methods is to determine the suitable step size for the policy update while an improper step size will either prolong the policy searching process or even degrade the policy, since the sampling data highly depends on the current policy. As a result, the trade-off between learning speed and robustness is a challenging problem for PG methods.\\

The trust region policy optimization (TRPO) method \cite{Ref9} proposes an objective function (also called the "surrogate" objective) and aims to maximize the objective function while subjecting to a constraint on the size of the policy update by KL divergence. However, the subjecting item is computationally inefficient and difficult to scale up for high-dimensional problems when extending to complex network architectures. To address this problem, the Proximal Policy Optimization (PPO), which adopts a clipping mechanism on the probability of likelihood ratio, was introduced \cite{Ref11}. This clipping mechanism significantly reduces the complexity of the optimizing process. To deal with learning robustness PPO tries to remove the incentive for pushing the policy away from the old one when the ratio is out of a clipping range, being more efficient for both implementation and computation than the trust region-based TRPO.\\
Despite the success of PPO it also has some flaws in restricting the update step size, which means that PPO can not truly bind the likelihood ratio within the clipping range \cite{Ref12}. Aiming towards a better step size update, in 2019 PPORB applied a rollback operation on the clipping function to inhibit the policy from being pushed away during training \cite{Ref13}. PPORB adopts a straight downward-slope function instead of the original flat function when the ratio is out of the clipping range, which hinders the natural incentive from PPO to seek a large policy update. However, this solution introduces new problems, such as when the ratio is extremely big and the clipped ratio will shoot to infinity or negative infinity, and thus contrast with its original aim. Additionally, the innerworkings of PPORB's hyperparameter $\alpha$, the slope of the rollback function, is still abstruse and poorly understood.\\
Inspired by the insights above, we propose a novel PPO clipping method, named Proximal Policy Optimization Smoothed algorithm (PPOS) which combines the strengths of both PPO and PPORB. We apply a functional clipping mechanism to prevent large policy updates while keeping the convergence of the clipped ratio. We compare PPOS with PPO and PPORB in high-dimensional control environments. We also analyze the hyperparameter here introduced in relation to the dimensions of five problems and provide a useful guideline for readers to orient their hyperparametric choices according to the dimension of their problems.

\section{Preliminaries}
This section discusses the background information needed to understand our policy search approach. We first describe the reinforcement learning (RL) framework, which is based on the Markov decision process (MDP). Then, we give an overview of policy search methods. We continue by introducing trust-region methods and actor-critic methods. Finally, we delve into the Proximal Policy Optimization (PPO) algorithm, which is currently one of the state-of-the-art algorithms in policy search.\\
A RL procedure consists of an agent and an environment, where the agent can interact with the environment and gain rewards during this interaction. We consider RL as a process running by a Markov decision process (MDP). An MDP is defined by the tuple $\left(\mathcal{S}, \mathcal{A}, p, p_{0}, r , \gamma\right)$, where $\mathcal{S}\subseteq \mathbb{R}^{d_{s}}$ is a finite set of states of the agent and $\mathcal{A} \subseteq \mathbb{R}^{d_{A}}$ is a finite set of actions which can be conducted by the agent. $p\left(s_{t+1} \mid s_{t}, a_{t}\right)$ denotes the probability of transfer from state $s_{t}$ to $s_{t+1}$ given the action conducted by the agent at time step $t$. $p\left(s_{t+1} \mid s_{t}, a_{t}\right)$ is usually stable and unknown in high-dimensional control tasks, but we can sample from $p\left(s_{t+1} \mid s_{t}, a_{t}\right)$ in simulation or from a physical system in the real world. $p_{0}$ denotes the initial state distribution of the agent, and $r(s_{t},a_{t})$ denotes the valued reward in state $s_{t}$ when the agent executes action $a_{t}$. The last $\gamma \in(0,1)$ is the discount factor. A policy in RL is defined as the map (or function) from the state space to the action space. The goal of RL is to find a stationary policy $ \pi (a_{t} \mid s_{t})$ that maximizes the long-term expected reward \cite{Ref9},
\begin{equation}\max_{\pi}\quad\mathbb{E}_{p_{\pi}(s) \pi(a \mid s) }\left[Q^{\pi}(s, a)\right]\label{eq1}\end{equation}
where\begin{equation}
   Q^{\pi}(s_{t}, a_{t})=\mathbb{E}_{s_{t+1}, a_{t+1}, \ldots}\left[\sum_{t=0}^{T} \gamma^{t} r\left(s_{t}, a_{t}\right)\right]
\end{equation}
in which $s_{0} \sim p_{0}(s), s_{t+1} \sim p\left(s_{t+1} \mid s_{t}, a_{t}\right)$, $a_{t} \sim \pi\left(a_{t} \mid s_{t}\right)$ and T is the terminal time step. $Q^{\pi}(s_{t}, a_{t})$ is called state-action value function which evaluate the state-action pair. We could also compute the value function $V^{\pi}\left(s_{t}\right)=\mathbb{E}_{a_{t}, s_{t+1}, \ldots}\left[\sum_{t=0}^{T} \gamma^{t} r\left(s_{t}, a_{t}\right)\right]$ which is the evaluation of the state $s_{t}$, and advantage function $A^{\pi}\left(s_{t}, a_{t}\right)=Q^{\pi}\left(s_{t}, a_{t}\right)-V^{\pi}\left(s_{t}\right)$ which indicates how reasonable the chosen action $a_{t}$ is given the state $s_{t}$.\\
In policy search, the goal is to find a policy $\pi$ that maximized Eq.\ref{eq1}. The performance of a policy $\pi$ is defined as $\eta(\pi)=\mathbb{E}_{s \sim \rho^{\pi}, a \sim \pi}[c(s, a)]$ where $\rho^{\pi}(s)=(1-\gamma) \sum_{t=0}^{T} \gamma^{t} \rho_{t+1}^{\pi}(s), \rho_{t+1}^{\pi}$ is the density function of state at time $t+1$. We could parameterize policies $\pi_{\theta}(a \mid s)$ with parameter vector $\theta$. In recent work, the policy usually adopts the neural network structure (e.g. \cite{Ref1} , \cite{Ref18} and \cite{Ref19}) which simulates the 
network of neuron of human-being to some extent. A policy can be improved by update the policy by the following surrogate performance objective which is called policy gradients methods \cite{Ref14},
\begin{equation}L_{\pi_{\mathrm{old}}}^{\mathrm{PG}}(\pi)=\mathbb{E}_{s, a}\left[r_{s, a}(\pi) A^{\pi_{\mathrm{old}}}(s, a)\right]+\eta\left(\pi_{\mathrm{old}}\right)\end{equation}
where $r_{s, a}(\pi)=\pi(a \mid s) / \pi_{\mathrm{old}}(a \mid s)$ is the likelihood ratio between the new policy $\pi$ and the old policy $\pi_{old}$, $A^{\pi_{\mathrm{old}}}(s, a)$ is the advantage function of the old policy $\pi_{old}$.\\
\subsection{Trust Region Policy Optimization}
The common way of conducting policy updates is using the samples from the old policy $\pi_{old}$ to improve the current policy $\pi$ at each iteration. However, the update may not be valid for the new policy in practice because the estimates for $p_{\pi}(\boldsymbol{s})$ and $Q^{\pi}(s, a)$ are based on the old policy which could result in some extremely large updates. To address this problem, Trust-Region Optimization (TRO) methods are used for keeping the new policy not far away from the old policy, which was first introduced in the relative entropy policy search (REPS) algorithm \cite{Ref10}. Many variants of TRO methods were introduced after the REPS algorithm (e.g.\cite{Ref15},\cite{Ref16},\cite{Ref17}). Most of these TRO methods use a bound of the KL-divergence of the policy update which prevents the policy update from being unstable while keeping the update suitable. Trust region policy optimization (TRPO) \cite{Ref9} adopts this bound to optimize the policy. The surrogate objective of TRPO is maximized while being subjected to a constraint on the KL-divergence between the old and current policy,
\begin{equation} \max_{\pi} \quad \mathbb{E}_{s, a}\left[\frac{\pi(a \mid s)} { \pi_{\mathrm{old}}(a \mid s)} A^{\pi_{\mathrm{old}}}(s, a)\right]\end{equation}
\begin{equation}\text { subject to } \quad {\mathbb{E}}_{s,a}\left[\operatorname{KL}\left[\pi_{\text {old }}\left(\cdot \mid s\right), \pi\left(\cdot \mid s\right)\right]\right] \leq \delta \end{equation}
where $\delta$ is a constant hyper-parameter and “·” means any action. In general, when $\delta$ is small enough, the update is valid since the difference between the new policy and the old policy is not substantial.
\subsection{Proximal Policy Optimization}
TRPO has been theoretically instantiated as capable of ensuring monotonic performance improvements of the policy. Nonetheless, due to the KL-divergence constraint TRPO is computationally inefficient and difficult to scale up for high-dimensional problems. In order to simplify the implementation and computational requirements, the PPO algorithm \cite{Ref11} was introduced. PPO restricts the policy update by a clipping function $\mathcal{F}$ on the ratio $r_{s, a}(\pi)$ directly,
\begin{equation}L^{\mathrm{CLIP}}(\pi)=\mathbb{E}\left[\min \left(r_{s, a}(\pi) A_{s, a}, \mathcal{F}\left(r_{s, a}(\pi), \epsilon\right) A_{s, a}\right)\right]\label{eq6}\end{equation}
In PPO, $\mathcal{F}$ is defined as,
\begin{equation}
\mathcal{F}^{\mathrm{PPO}}\left(r_{s, a}(\pi), \epsilon,\alpha\right)=\left\{\begin{array}{ll}
1-\epsilon & r_{s, a}(\pi) \leq 1-\epsilon \\
1+\epsilon  & r_{s, a}(\pi) \geq 1+\epsilon \\
r_{s, a}(\pi) & \text { otherwise }
\end{array}\right. 
\label{eq7}
\end{equation}
where $1-\epsilon$ and $1+\epsilon$ are called the lower and upper clipping range, respectively, and $\epsilon \in (0,1)$ is a hyper parameter. The minimization operation in Eq. (\ref{eq6}) is to keep the gradient when the $r_{s, a}$ is out of the clipping range. Multiplied by $A_{s, a}$ the illustration of the clipping mechanism as the red dash line can be seem in Fig.\ref{ClipFunction}. \\
Both TRPO and PPO adopt the actor-critic method \cite{Ref20}, in which the Value function $Q^{\pi}(s, a)$ is estimated by a parameterized function approximator $\widehat{Q}(s, a; \omega)$, where the vector $\omega$ is the parameter of the approximator e.g. a linear model or usually a neural network. This estimator is called the critic which can have a lower variance than traditional Monte Carlo estimators \cite{Ref26}. The critic's estimate is used to optimize the policy $\pi(a \mid s ; \theta)$ which is also called the actor parameterized by $\theta$. The actor is usually represented by a neural network. After that, the policy parameters can be updated by the gradient of surrogate objective Eq. (\ref{eq6}) as,
\begin{equation}
    \theta_{t+1}= \theta_{t}+\beta \nabla \hat{L}^{\mathrm{CLIP}}\left(\theta_{t}\right)
    \label{eq8}
\end{equation}where $\beta$ is the step size and $\nabla \hat{L}^{\mathrm{CLIP}}\left(\theta_{t}\right)$ denote the gradient of $\hat{L}^{\mathrm{CLIP}}$ at $\theta_{t}$. \\
\subsection{PPO with Rollback}
\label{2.3}
PPO has made progress in different control tasks, especially with high-dimensional observation and action space, but it also has raised concerns of researchers about whether this clipping mechanism can truly restrict the policy (\cite{Ref21} \cite{Ref12}). In reality, PPO is uncapable of binding the likelihood ratio within the clipping range because its clipping mechanism still pushes the ratio out of its range, rendering the policy out of the bounds.\\  
To address this problem, PPORB \cite{Ref13} proposed a rollback operation on the clipping function attempting to remove the incentive of moving far away, which is defined as 
\begin{equation}\mathcal{F}^{\mathrm{PPORB}}\left(r_{s, a}(\pi), \epsilon, \alpha\right)=\left\{\begin{array}{ll}
-\alpha r_{s, a}(\pi)+(1+\alpha)(1-\epsilon) & r_{s, a}(\pi) \leq 1-\epsilon \\
-\alpha r_{s, a}(\pi)+(1+\alpha)(1+\epsilon) & r_{s, a}(\pi) \geq 1+\epsilon \\
r_{s, a}(\pi) & \text { otherwise }
\end{array}\right.\label{eq9}\end{equation}
where $\alpha$ is a hyperparameter to decide the force of the rollback operation.
When the ratio $r_{s, a}$ is out of the clipping range, it will make negative effects on the output instead of keeping the value as a constant. PPORB has made vast progress from the original PPO algorithm, but at the same time it leaves space for improvement. Firstly, the rollback operation can not converge when the ratio is extremely large: In high dimensional tasks, it will make the policy search sway around the optimal policy. Secondly, the rollback operation introduces an extra hyperparameter which can be only tuned by experience with that same problem. However, the hyperparametric possibilities result in significantly different results, with the values for this parameter ranging from 0.02 to 0.3.


\section{Method}

In this section, we first describe our reinforcement learning (RL) method, and then compare it with other methods of other proximal policy optimization algorithms. Finally, we discuss and prove the stability and strength of our method in this paper. 
\subsection{Proximal Policy Optimization Smoothed Algorithm}
\label{sec3.1}
As discussed in the section Preliminaries \ref{2.3}, while the rollback operation can, to some extent, restrict the likelihood ratio, it also leaves unsolved problems: the convergence property and the choice of hyperparameters. We deploy our method on the structure of PPO with a general form for sample (s,a), 
\begin{equation}L_{s, a}(\pi)=\min \left(r_{s, a}(\pi) A_{s, a}, \mathcal{F}\left(r_{s, a}(\pi), \cdot\right) A_{s, a}\right)\end{equation}
where $\mathcal{F}$ is a clipping function which attempts to restrict the policy update and “·” means any hyperparameters of $\mathcal{F}$. In PPO, $\mathcal{F}$ is a ratio-based clipping function $\mathcal{F}^{\mathrm{PPO}}$ as Eq. (\ref{eq7}), which will be flat whenever out of the clipping range. In PPORB, $\mathcal{F}^{\mathrm{PPORB}}$ has the rollback operation as Eq. (\ref{eq9}) instead of keeping it flat. We modify this function to promote the performance and stability of restricting the likelihood ratio by a functional clipping method, which is defined as
\begin{equation}
\mathcal{F}^{\mathrm{PPOS}}\left(r_{s, a}(\pi), \epsilon,\alpha\right)=\left\{\begin{array}{ll}
-\alpha tanh(r_{s, a}(\pi)-1)+1+\epsilon+\alpha tanh(\epsilon) & r_{s, a}(\pi) \leq 1-\epsilon \\
-\alpha tanh(r_{s, a}(\pi)-1)+1-\epsilon-\alpha tanh(\epsilon) & r_{s, a}(\pi) \geq 1+\epsilon \\
r_{s, a}(\pi) & \text { otherwise }
\end{array}\right. 
\end{equation}
\begin{figure}[htbp]
\centering
\subfigure[A$>$0]{
\includegraphics[width=6.5cm]{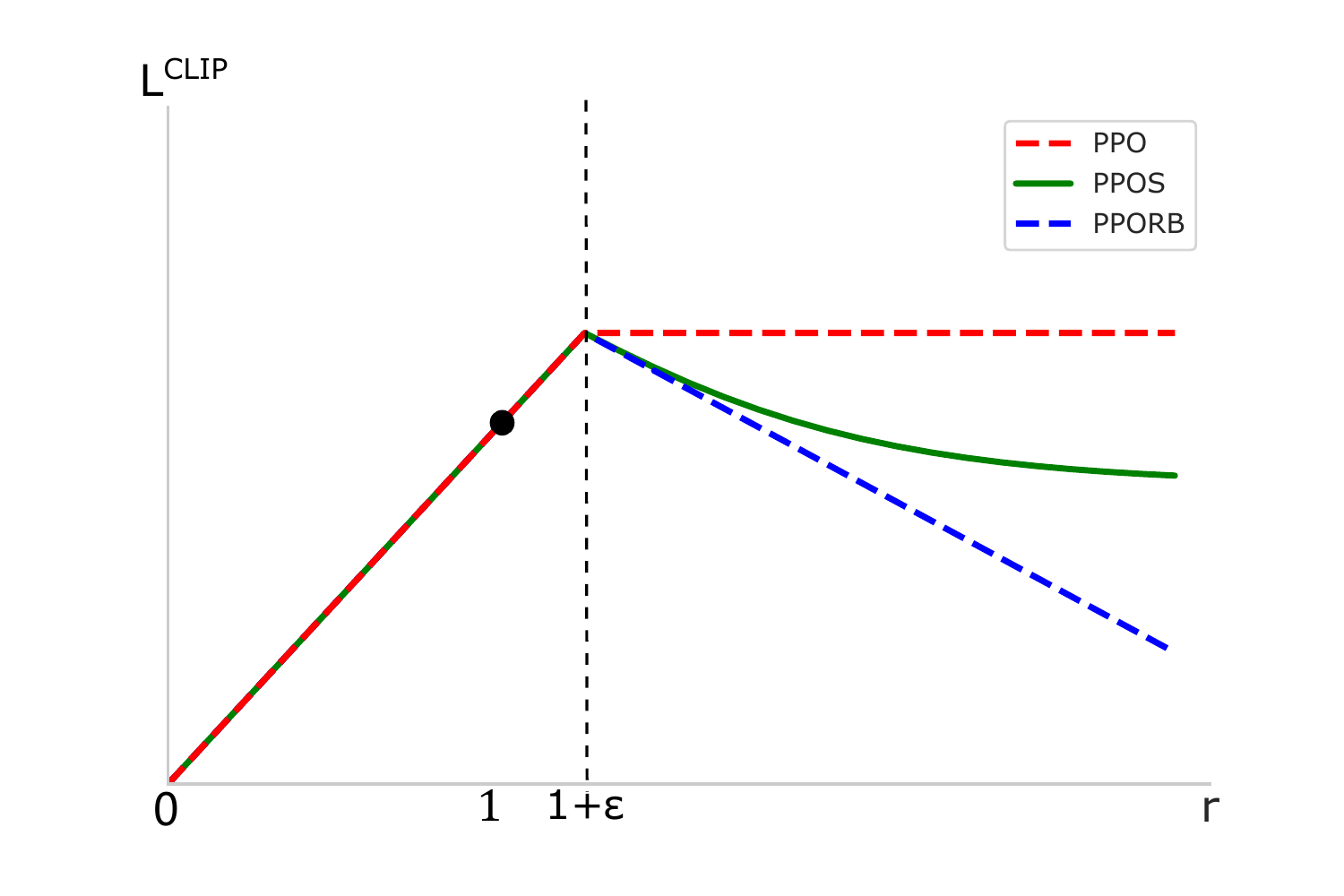}
}
\quad
\subfigure[A$<$0]{
\includegraphics[width=6.5cm]{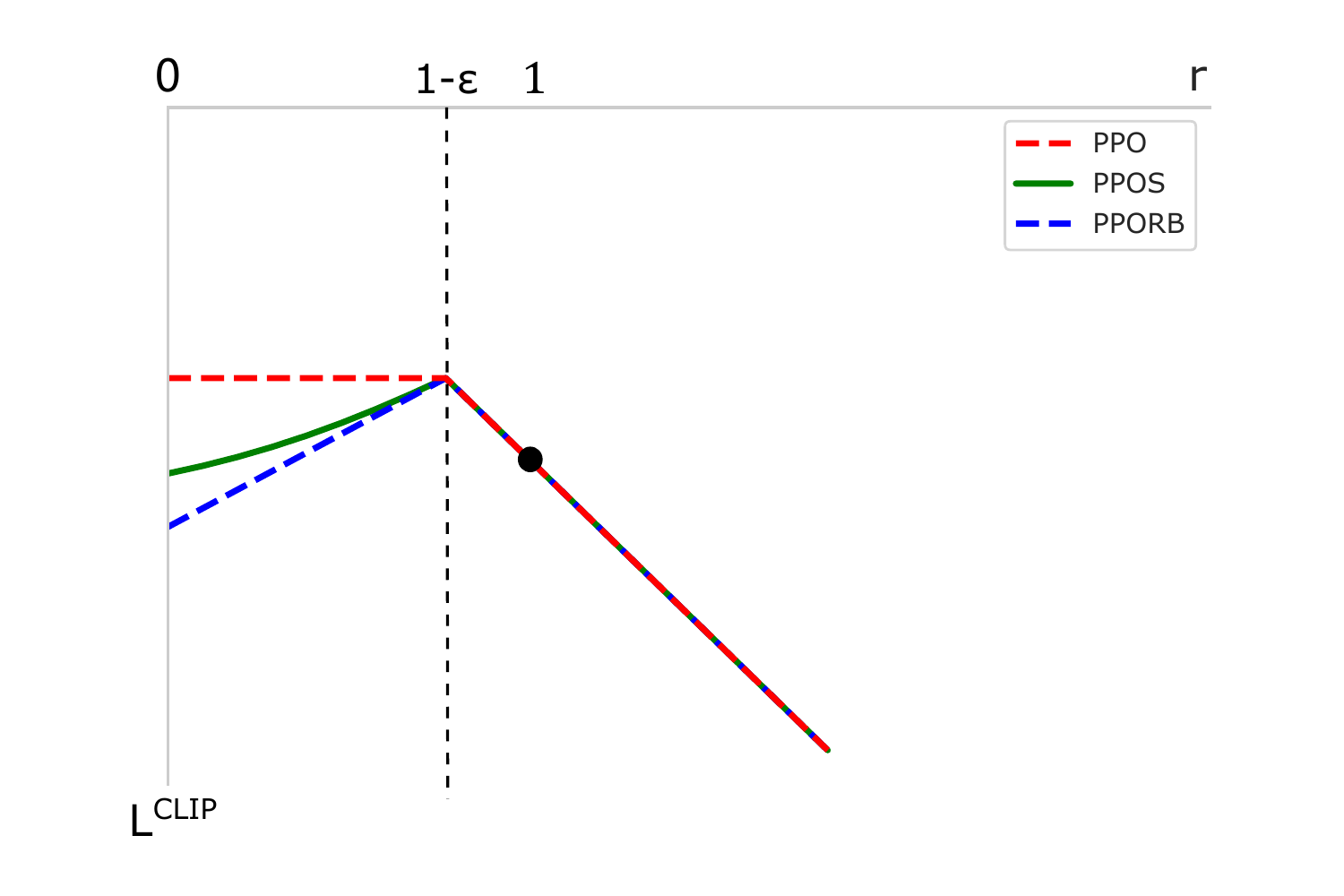}
}
\caption{Plots showing the clip function $L^{CLIP} $ of the likelihood ratio $r_{s, a}$ of PPO and PPOS, for positive advantages (left) and negative advantages (right). The black circle on each plot shows the starting point for the optimization, i.e., r = 1. When r exceeds the clipping range, the slope of $L^{CLIP} $ of PPOS goes along with the tanh function, while the slope of $L^{CLIP} $ of PPO is flattened.}
\label{ClipFunction}
\end{figure}
where $\alpha$ is a hyperparameter to decide the scale of the functional clipping. The decreasing slope will be more steep when $\alpha$ is larger.\\
Combined the original PPO algorithm with the novel function clipping method, we obtain the Proximal Policy Optimization Smoothed algorithm as shown in Fig. \ref{ClipFunction} which plots $L^\mathrm{CLIP}(\pi)$ as functions of the likelihood ratio $r_{s, a}(\pi)$. We utilize the gradient change of hyperbolic tangent activation function to obtain the smoothed effect. As the figure depicted, when  $r_{s, a}(\pi)$ is out of the clipping range, the slope of $L^\mathrm{PPOS}(\pi)$ starts with a large absolute value and generally converges to zero. This smoothing mechanism guarantees the convergence of the clipping function. When $\alpha<0$, the function is more natural and smoother in shape. However, the actual performance will not be better than $\alpha>0$ which can be viewed in section \ref{sec4.3}. At the same time, the hyperbolic tangent activation function could be replaced by other functions to explore more.\\
In contrast to PPORB, PPOS will never have a negative infinite output which is unacceptable in policy search. Formally, we obtain and prove the following theorem where we utilize functions of $\pi$ and subscript t to represent the functions of sample $(s_t,a_t)$ for simplicity, e.g., $r_{t}(\theta) \triangleq r_{s_{t}, a_{t}}\left(\pi_{\theta}\right)$ and $L_{t}(\theta) \triangleq L_{s_{t}, a_{t}}\left(\pi_{\theta}\right)$.\\
\subsection{Theorem}
In here we elaborate on the mathematical reasoning behind PPOS, and show the mechanism behind our better ratio restriction than PPO and PPORB. Given the parameter $\theta_{0}$, let $\theta_{1}^{\mathrm{PPORB}}=\theta_{0}+\beta \nabla \hat{L}^{\mathrm{PPORB}}\left(\theta_{0}\right), \theta_{1}^{\mathrm{PPOS}}=\theta_{0}+\beta \nabla \hat{L}^{\mathrm{PPOS}}\left(\theta_{0}\right)$.
The set of indexes of the samples which satisfy the clipping condition is denoted as 
\begin{equation}
  \Omega=\left\{t|1 \leq t \leq T,| r_{t}\left(\theta_{0}\right)-1 \mid \geq \epsilon \text { and } r_{t}\left(\theta_{0}\right) A_{t} \geq r_{t}\left(\theta_{\text {old }}\right) A_{t}\right\}
\end{equation}

If $t\in \Omega$ and $r_t(\theta_{0})$ satisfies $\sum_{t^{\prime} \in \Omega}\left\langle\nabla r_{t}\left(\theta_{0}\right), \nabla r_{t^{\prime}}\left(\theta_{0}\right)\right\rangle A_{t} A_{t^{\prime}}>0$, then there exists a $\beta^*$ such that for any $\beta\in(0,\beta^{*})$, we have 
\begin{equation}
  \left|r_{t}\left(\theta_{1}^{\mathrm{PPOS}}\right)-1\right|<\left|r_{t}\left(\theta_{1}^{\mathrm{PPORB}}\right)-1\right|
\end{equation}
$\mathrm{Proof}$:\\
Let $\phi(\beta)=r_{t}\left(\theta_{0}+\beta \nabla \hat{L}^{\mathrm{PPOS}}\left(\theta_{0}\right)\right)-r_{t}\left(\theta_{0}+\beta \nabla \hat{L}^{\mathrm{PPORB}}\left(\theta_{0}\right)\right)$.
By chain rule, we can have
\[
\begin{aligned}
\phi^{\prime}(0) &=\nabla r_{t}^{\top}\left(\theta_{0}\right)\left(\nabla \hat{L}^{\mathrm{PPOS}}\left(\theta_{0}\right)-\nabla \hat{L}^{\mathrm{PPORB}}\left(\theta_{0}\right)\right) \\
&=-\alpha \sum_{t^{\prime} \in \Omega}\left\langle\nabla r_{t}\left(\theta_{0}\right), \nabla r_{t^{\prime}}\left(\theta_{0}\right)\right\rangle A_{t^{\prime}}
\end{aligned}
\]
Because $r_{t}\left(\theta_{0}\right) \geq 1+\epsilon$ and $A_{t}>0,$ we have $\phi^{\prime}(0)<0$. Thus, there exists ${\beta^*}>0$ such that for any $\beta \in(0, {\beta^*})$
\[
\phi(\beta)<\phi(0)=0
\]
Hence, we have
\[
r_{t}\left(\theta_{1}^{\mathrm{PPOS}}\right)<r_{t}\left(\theta_{1}^{\mathrm{PPORB}}\right)
\]
We obtain
\[
\left|r_{t}\left(\theta_{1}^{\mathrm{PPOS}}\right)-1\right|<\left|r_{t}\left(\theta_{1}^{\mathrm{PPORB}}\right)-1\right|
\]
Similarly, for the case where $r_{t}\left(\theta_{0}\right) \leq 1-\epsilon$ and $A_{t}<0,$ we also have $\left|r_{t}\left(\theta_{1}^{\mathrm{PPOS}}\right)-1\right|<\left|r_{t}\left(\theta_{1}^{\mathrm{PPORB}}\right)-1\right|
$, which means that the PPOS function can choose a better ratio closer to 1. This way the proposed functional clipping mechanism can improve the ability in keeping the ratios within the range.


\subsection{Experimental setup}
In our Experiments section, we test the performance of our algorithm with the following benchmark: (a) PPO: the original PPO algorithm with the author recommended clipping range hyperparameter $\epsilon=0.2$, as seen in \cite{Ref11}. (b) PPORB: PPO with the rollback function. The rollback coefficient $\alpha$ is set to be 0.3 for all tasks (except for the Humanoid-v2 task) as suggested by the authors in \cite{Ref13} (c) PPOS: Our proposed method with an $\alpha$ coefficient set according to the Table \ref{tab:2}.\\
The algorithms are evaluated on continuous control benchmark tasks implemented on OpenAI Gym \cite{Ref24}, simulated by MuJoCo \cite{Ref23} in a Tianshou platform\footnote{https://github.com/thu-ml/tianshou}. The chosen MuJoCo environment is the version-2 (v2) for all experiments. The actor and critic networks are neural networks with one hidden layer and 128 neurons per layer, and the adopted activation function is ReLU function \cite{Ref27}. Each algorithm was run with ten random seeds and 250,000 timesteps, which are obtained by 2500 steps per epoch and 100 epochs (except for the Humanoid-v2 task with 1 million timesteps, due to its higher dimensionality) using ADAM optimizer \cite{Ref25}. The details can be viewed on the Table \ref{tab:3} in the Appendix.

\begin{figure}[htbp] \centering 
\subfigure{  
\includegraphics[width=2.4cm]{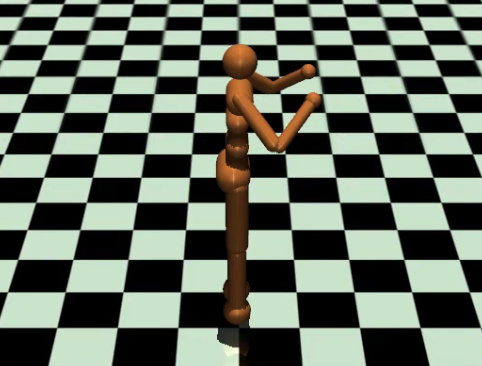} 
} 
\quad
\subfigure{ 
\includegraphics[width=2.4cm]{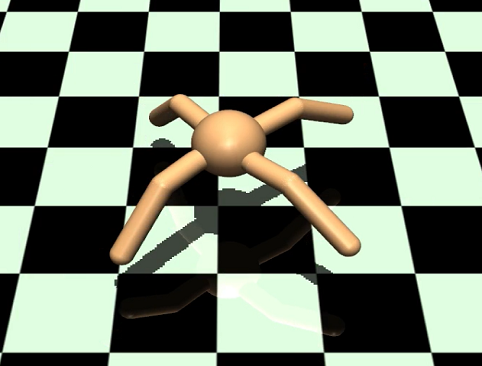} 
} 
\quad
\subfigure{  
\includegraphics[width=2.4cm]{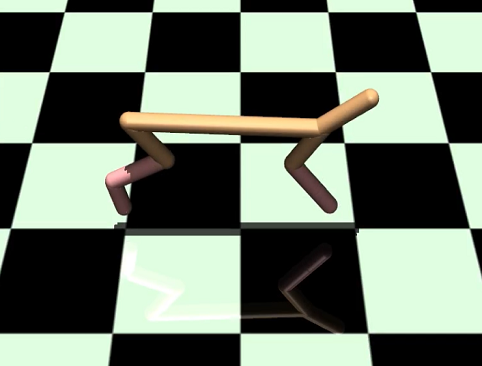} 
} 
\quad
\subfigure{  
\includegraphics[width=2.4cm]{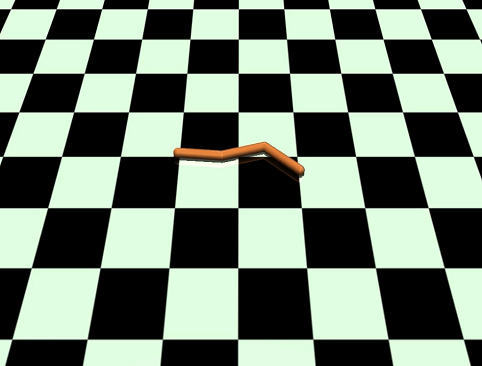} 
} 
\quad
\subfigure{  
\includegraphics[width=2.4cm]{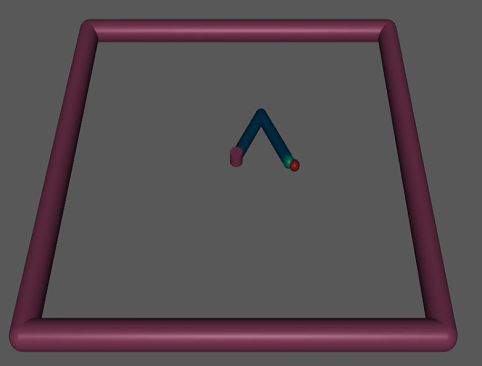} 
} 
\caption{Illustrations of five Mujoco tasks.Illustrations of five Mujoco tasks: Humanoid-v2, Ant-v2, HalfCheetah-v2, Swimmer-v2, and Reacher-v2.} 
\label{fig:2} 
\end{figure} 

\section{Experiments}
In this section, we present experiments to assess the performance of our proposed algorithm in terms of performance and stability during policy searches. We compare our method with aforementioned state-of-the-art PPO variants in different environments. We also take in consideration different hyperparametric choices and their corresponding results are shown, along with a quick guideline with tuning suggestions to quickly deploy our algorithm on different applications.

\subsection{Comparison with other methods}
\begin{figure}[htbp]
\centering
\subfigure[Humanoid-v2 $|\mathcal{O}|=376$]{
\includegraphics[width=6.5cm]{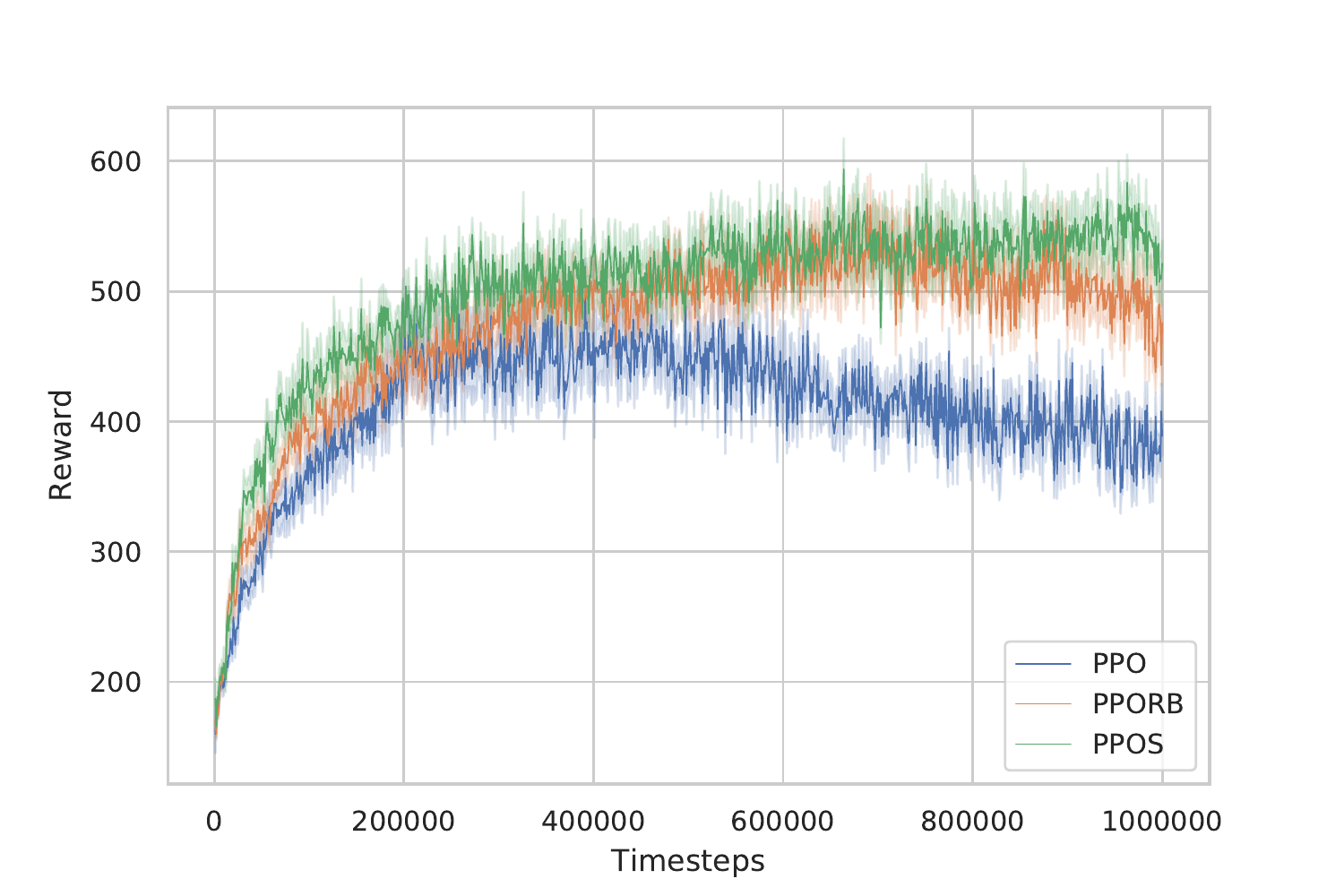}
}
\quad
\subfigure[Ant-v2 $|\mathcal{O}|=111$]{
\includegraphics[width=6.5cm]{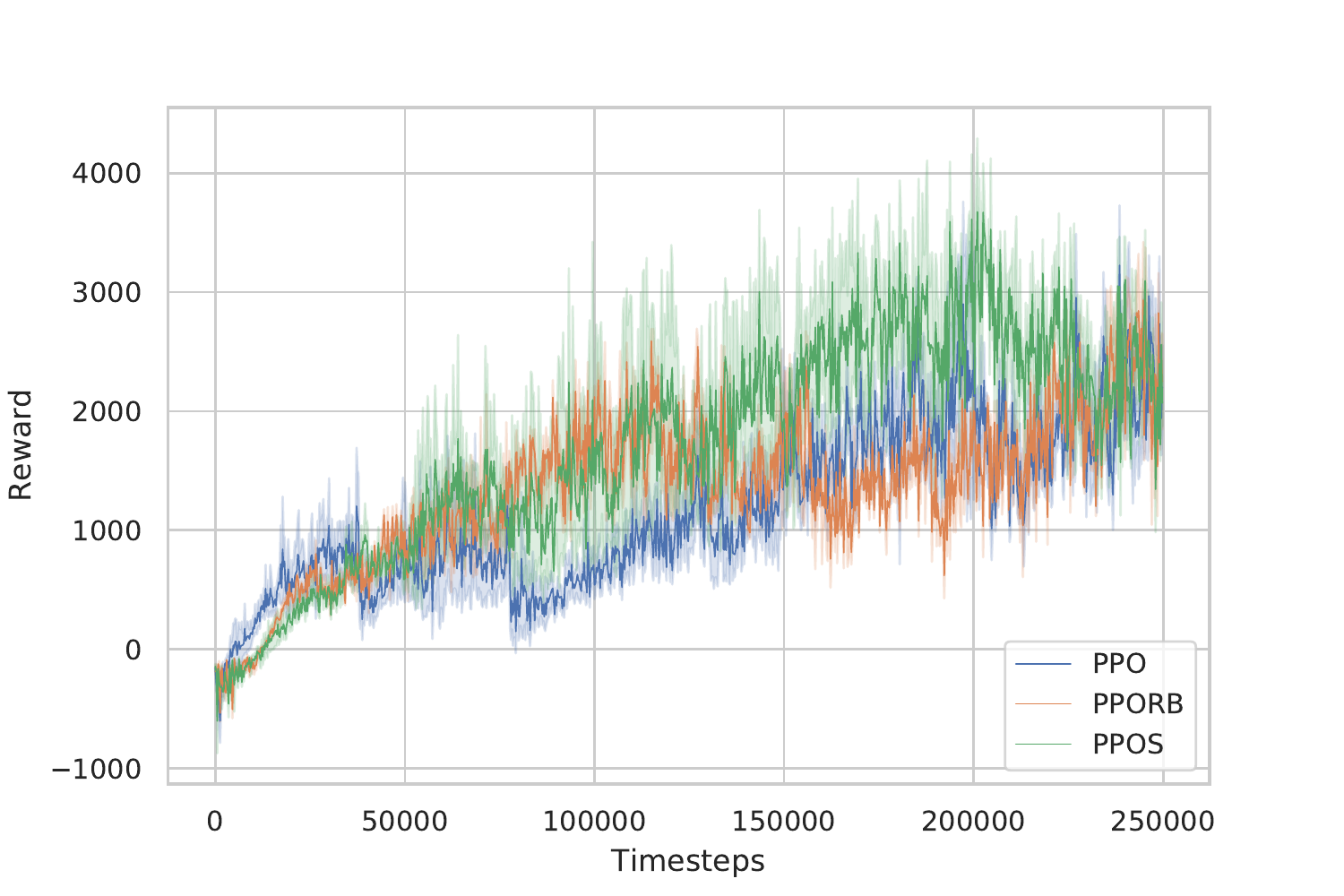}
}
\quad
\subfigure[HalfCheetah-v2 $|\mathcal{O}|=17$]{
\includegraphics[width=6.5cm]{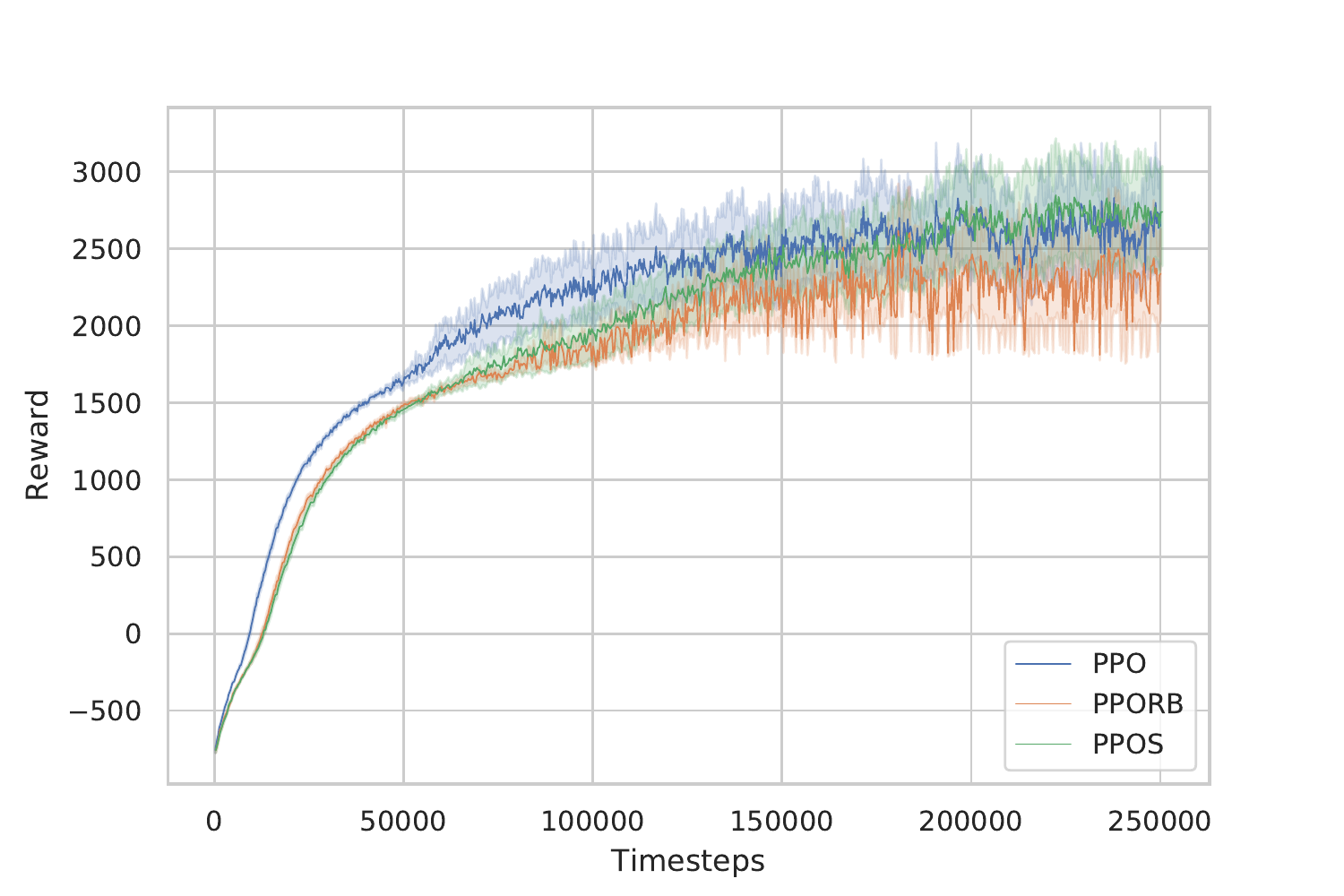}
}
\quad
\subfigure[Swimmer-v2 $|\mathcal{O}|=8$]{
\includegraphics[width=6.5cm]{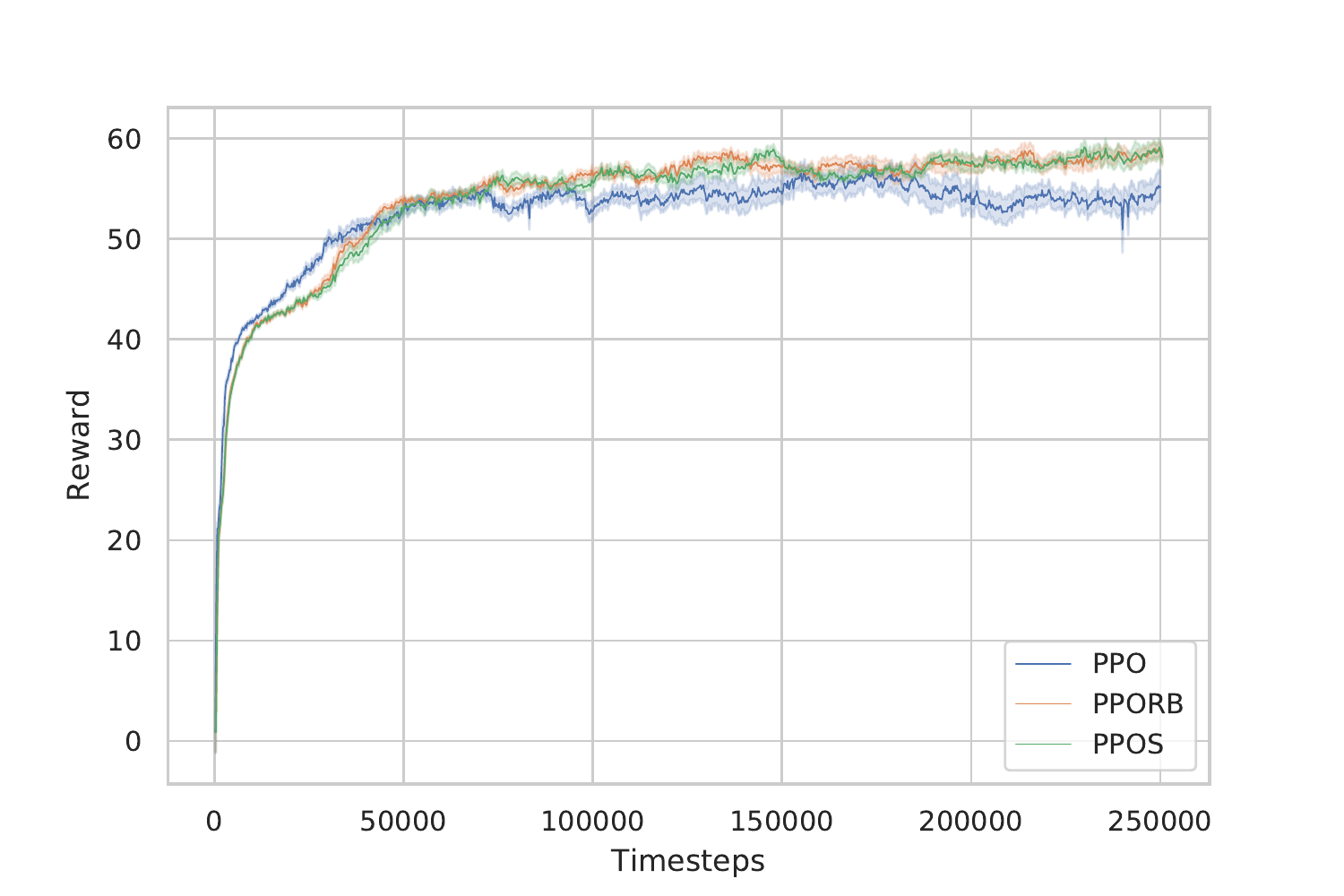}
}
\quad
\subfigure[Reacher-v2 $|\mathcal{O}|=11$]{
\includegraphics[width=6.5cm]{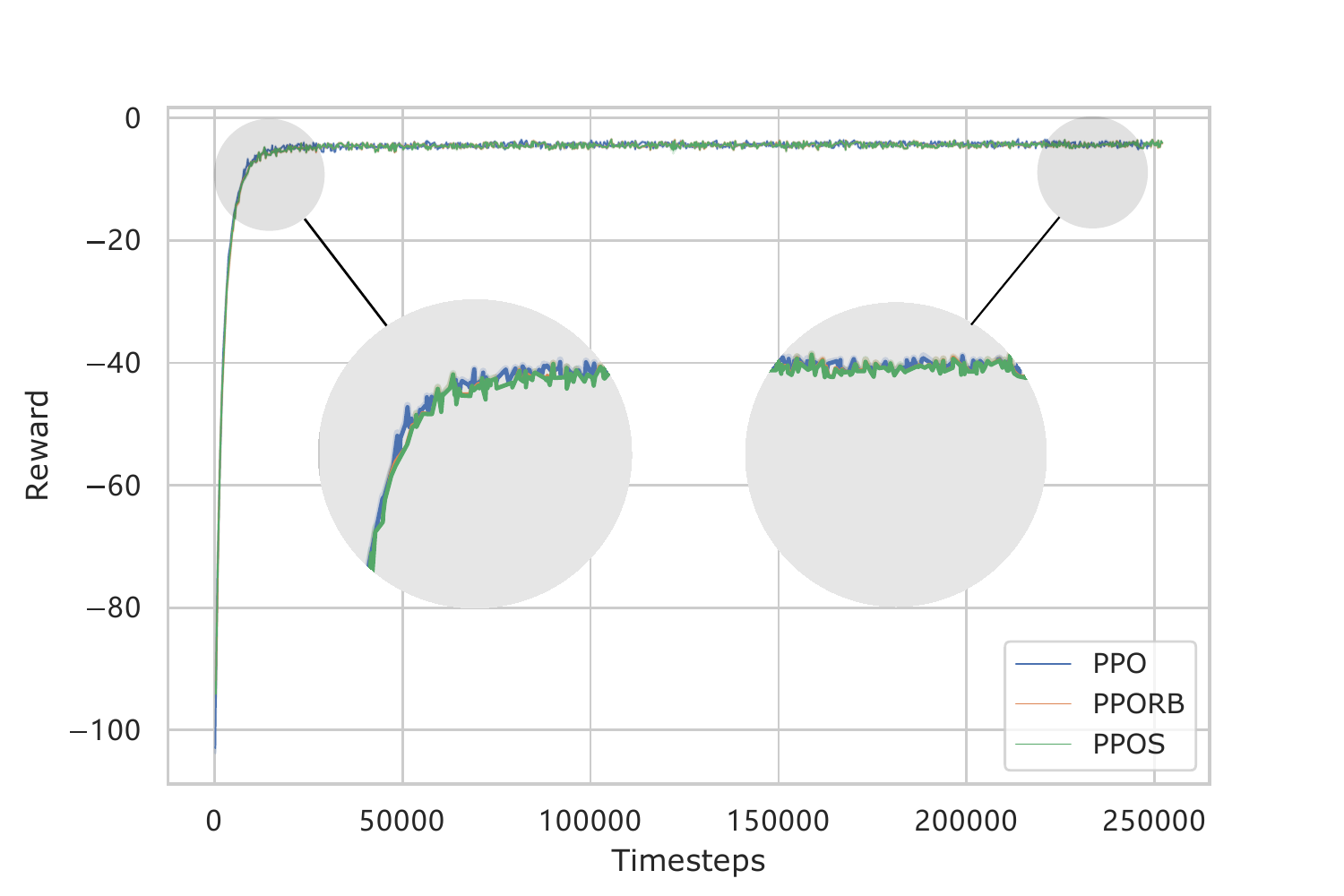}
}
\caption{Episode rewards of the policy during the training process averaged over ten random seeds on MuJoCo tasks, comparing our methods with PPO and PPORB. The solid lines are mean values of ten trials while the shaded area depicts the 60$\%$ confidence interval. $|\mathcal{O}|$ represents the dimension of observation.}
\label{fig:3}
\end{figure}

We conducted experiments in five Mujoco environments with continuous control tasks of both high and low dimensions, as shown in Fig. \ref{fig:2}. In Fig. \ref{fig:3} we plot episode rewards during the training process on the tasks. Our PPOS method with the functional clipping mechanism significantly outperforms both PPO and PPORB on hard tasks characterized by high observation and action dimension (e.g., Humanoid-v2 with $|\mathcal{O}|=376$ and $|\mathcal{A}|=17$ and Ant-v2 with $|\mathcal{O}|=111$ and $|\mathcal{A}|=8$), both in terms of learning speed and final rewards. For the HalfCheetah-v2 with medium dimensionality ($|\mathcal{O}|=17$ and $|\mathcal{A}|=6$) our method achieves a better final reward in the later stage. The advantage given by the clipping mechanism is less salient in lower dimensionality, and our PPOS is comparable to both PPO and PPORB in such tasks (e.g., Swimmer-v2 with $|\mathcal{O}|=8$ and $|\mathcal{A}|=2$ and Reacher-v2 with $|\mathcal{O}|=11$ and $|\mathcal{A}|=2$). The stricter clipping mechanism from both PPORB and PPOS have a tendency to slow down these algorithms on their initial stage (in special in low dimensions, while making it faster in higher dimensionalities), with smaller policy updates, but eventually it enables these algorithms to asymptote at higher rewards.\\
The numerical analysis with mean and standard deviation is presented on Table \ref{tab:1}. PPOS reaches a better performance in 80\% of the adopted Mujoco tasks, and, as seen on Table  \ref{tab:1}, it was virtually tied with PPORB on the low dimensional Swimmer-v2 task (and presented a lower standard deviation). On average, PPOS reaches 8\% higher rewards than PPO and 7\% higher than PPORB, while the standard deviation of PPOS is 3\% lower than both PPO and PPORB. A possible explanation for this is that the functional clipping mechanism successfully keeps the policy update within a reasonable range, while PPO and PPORB made the update either too large or too small. Besides, the nature of PPORB's clipping can generate very negative updates, which can be detrimental to the policy search.

\begin{table}[h]
\renewcommand\arraystretch{1.5}
\caption{Continuous control environments}
\label{tab:1}       
\begin{tabular}{p{4cm}|p{3cm}|p{3cm}|p{3cm}}
\hline\noalign{\smallskip}
Task & PPO & PPORB & PPOS  \\
\noalign{\smallskip}\hline\noalign{\smallskip}
Humanoid-v2 & $473.3\pm61.18$   & $499.4\pm59.15$  & \boldmath{$535.6\pm59.67$} \\
Ant-v2      &  $2481\pm380.6$   & $2582\pm404.9$   & \boldmath{$2964\pm189.3$} \\
HalfCheetah-v2 & $2625\pm164.5$ & $2341\pm399.7 $  &  \boldmath{$2684\pm191.1$}\\
Swimmer-v2  &  $53.68\pm5.126$  & \boldmath{$57.89\pm2.733$} &  $ 57.50\pm1.744 $ \\
Reacher-v2  &  $-4.306\pm0.053$ & $-4.294\pm0.061$ &  \boldmath{$-4.290\pm0.053 $} \\
\noalign{\smallskip}\hline
\end{tabular}
Mean of the average return over 50 iterations around the best reward $\pm$ standard deviation over ten random seeds of three methods. Bold denotes the best result with highest reward among three methods. 
\end{table}

\begin{figure}[htbp]
\centering
\subfigure[Humanoid-v2 $|\mathcal{O}|=376$]{
\includegraphics[width=6.5cm]{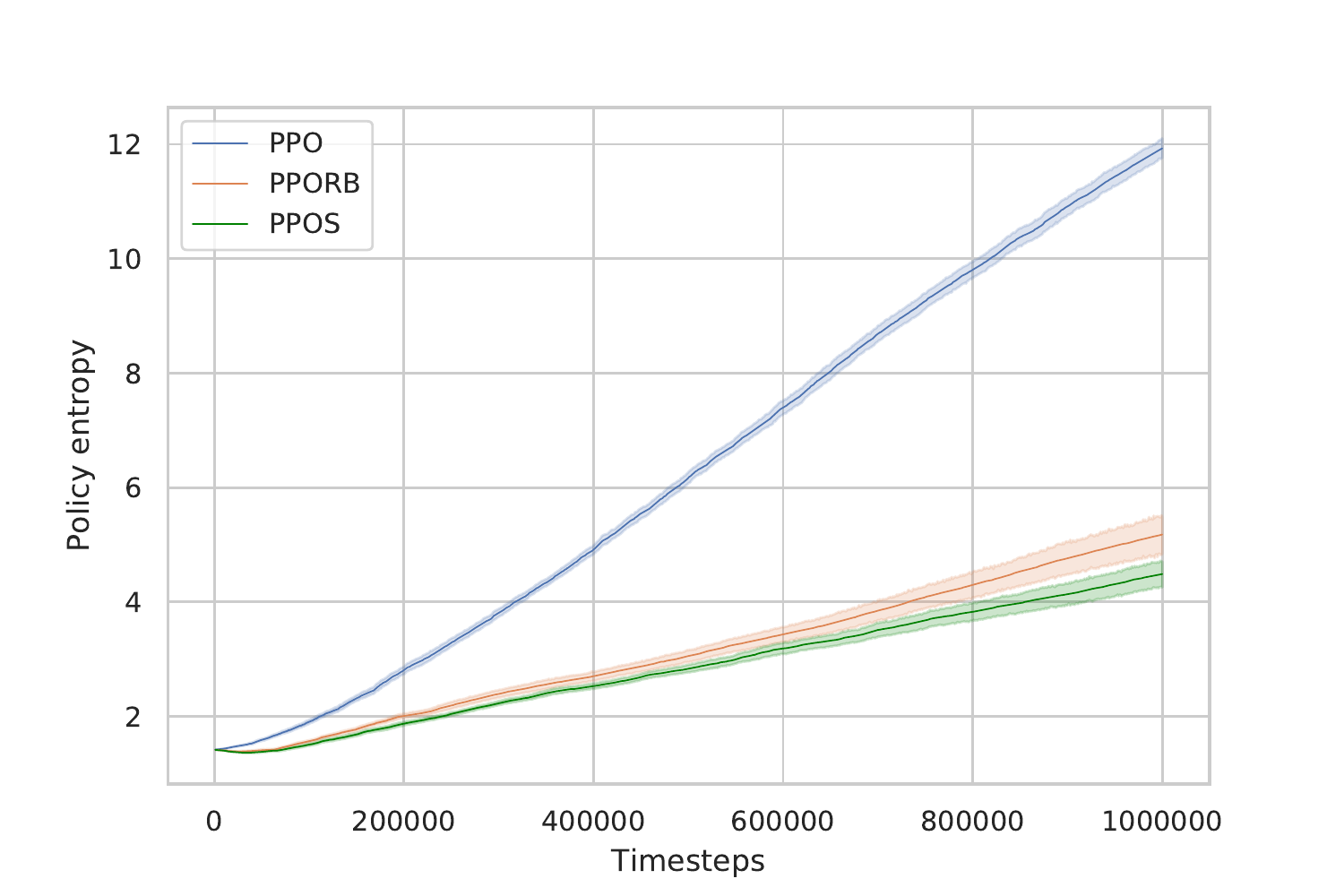}
}
\quad
\subfigure[Ant-v2 $|\mathcal{O}|=111$]{
\includegraphics[width=6.5cm]{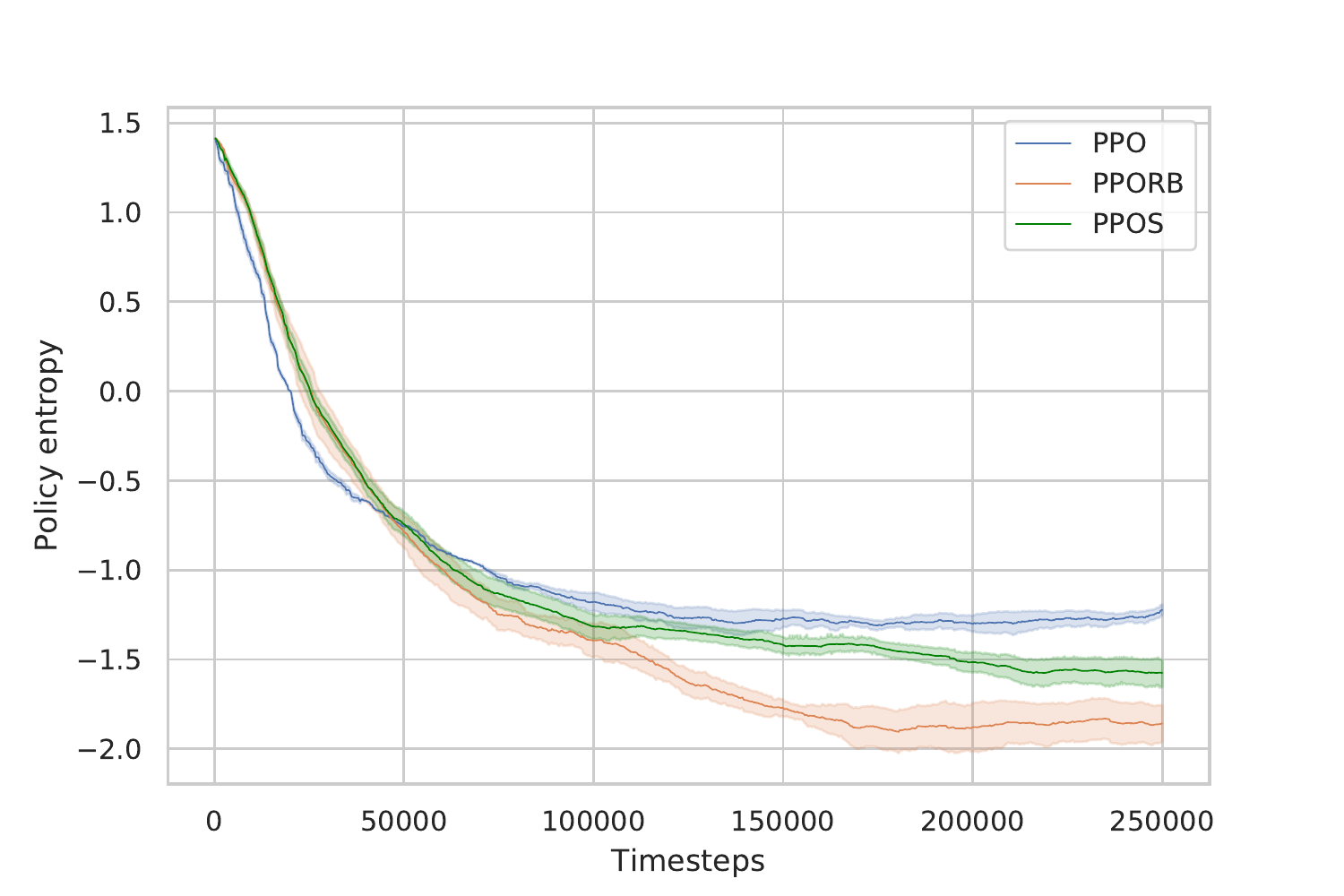}
}
\quad
\subfigure[HalfCheetah-v2 $|\mathcal{O}|=17$]{
\includegraphics[width=6.5cm]{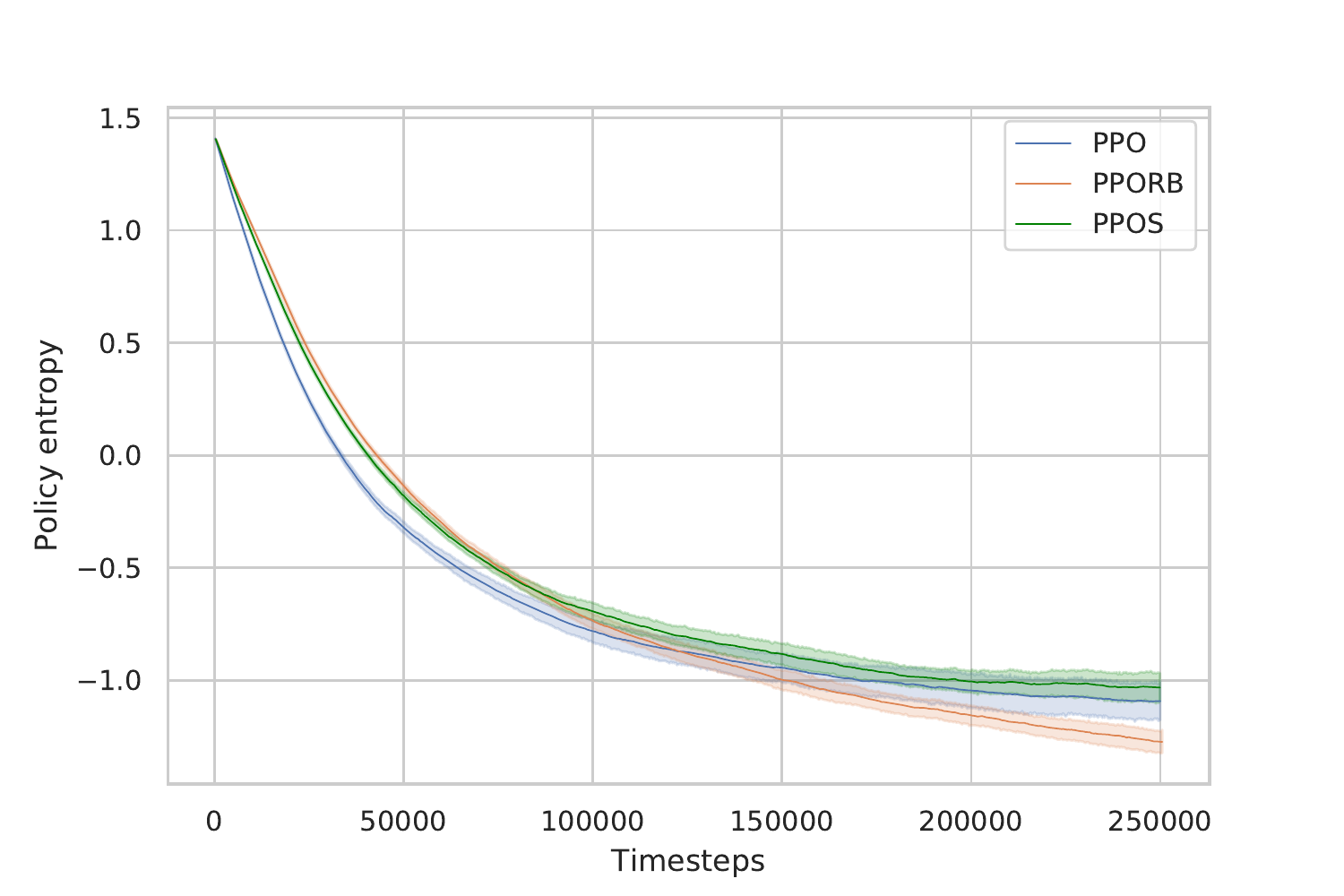}
}
\quad
\subfigure[Swimmer-v2 $|\mathcal{O}|=8$]{
\includegraphics[width=6.5cm]{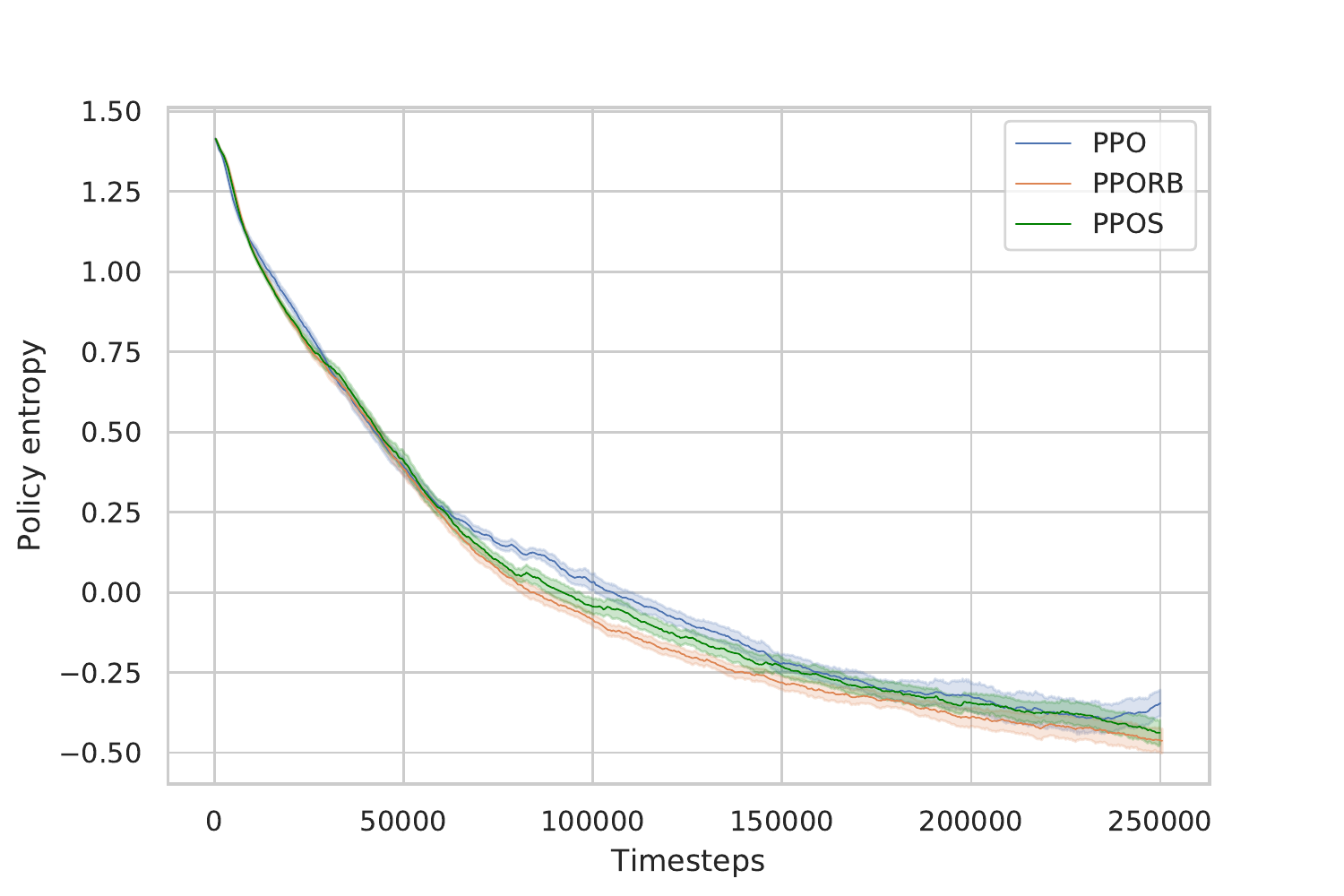}
}
\quad
\subfigure[Reacher-v2 $|\mathcal{O}|=11$]{
\includegraphics[width=6.5cm]{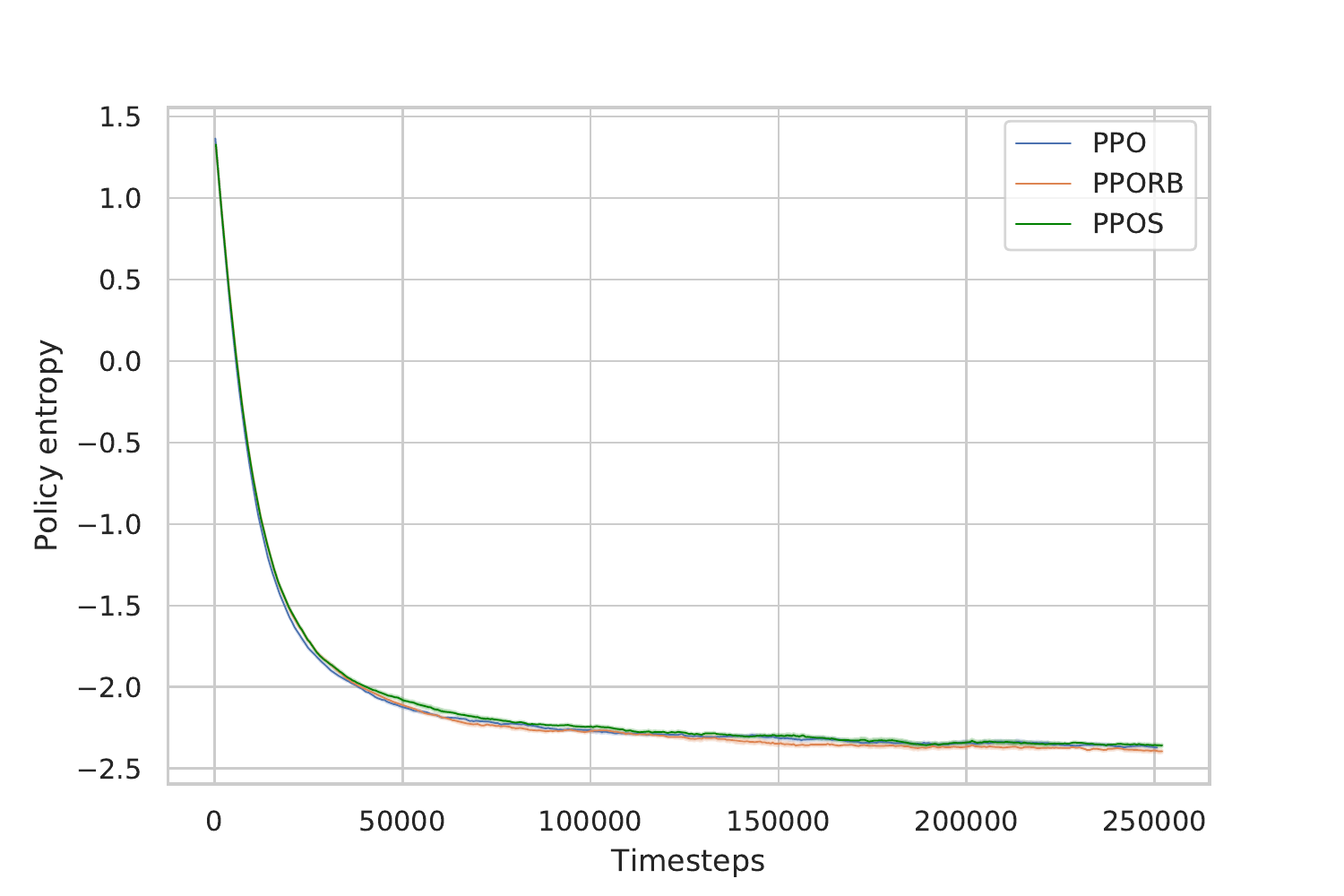}
}
\caption{Entropy of the policy during the training process, averaged over ten random seeds on MuJoCo tasks, in comparison with PPO and PPORB. The solid lines are mean values of ten trials while the shaded area depicts the 60$\%$ confidence interval. $|\mathcal{O}|$ represents the dimension of observation.}
\label{fig:4}
\end{figure}
We also compared three methods on entropy during all training procedures as shown in Fig. \ref{fig:4}. In general, PPOS and PPORB have lower entropy than PPO in most of the tasks which means PPOS and PPORB can make the trained policy more deterministic. In Humanoid-v2, PPOS can even have lower entropy than PPORB which means the policy is more practical for implements on real robots.

\subsection{Choice of the hyperparameter}
\label{sec4.3}
\begin{figure}
\centering
\subfigure[Humanoid-v2 $|\mathcal{O}|=376$]{
\includegraphics[width=6.5cm]{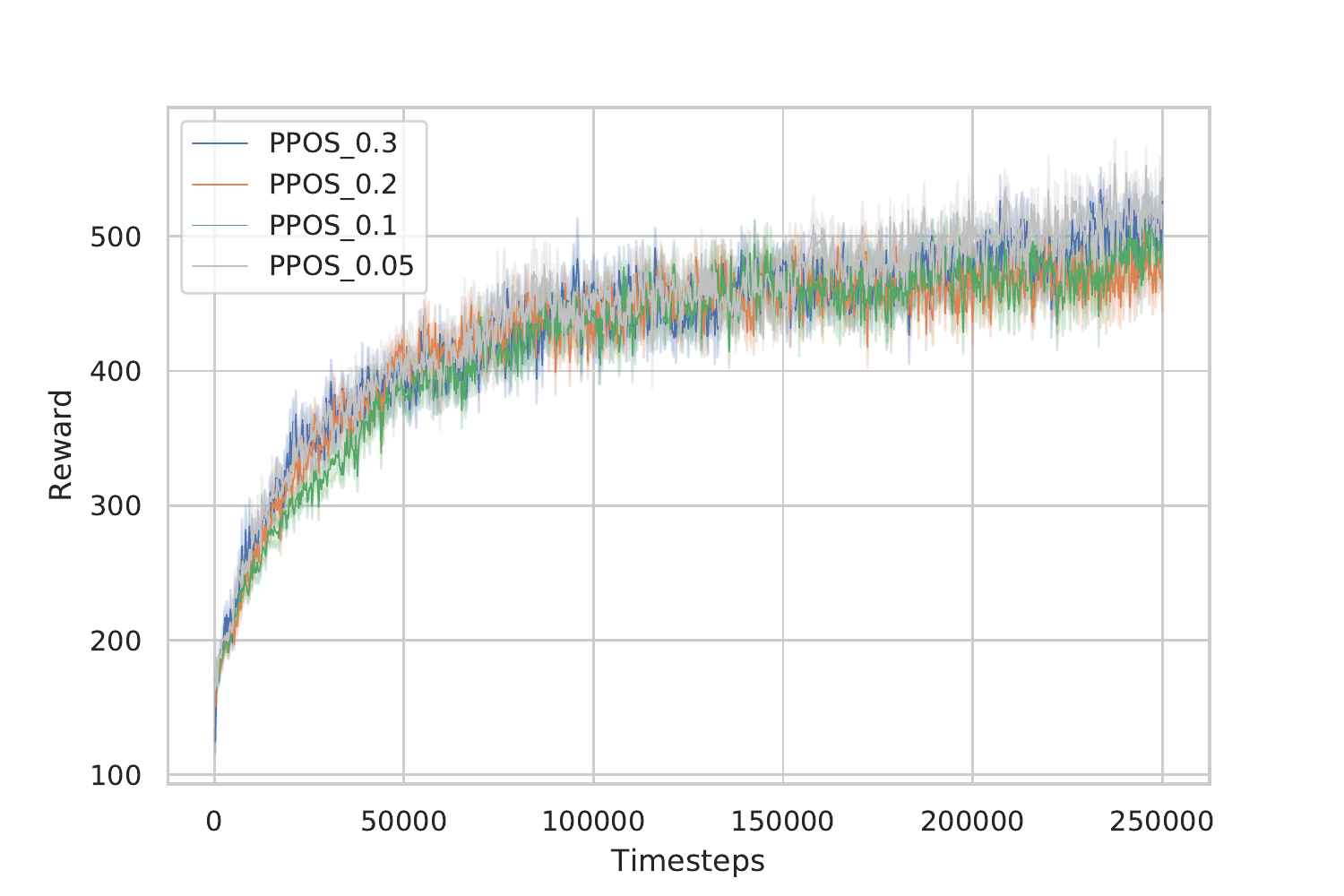}
}
\quad
\subfigure[Ant-v2 $|\mathcal{O}|=111$]{
\includegraphics[width=6.5cm]{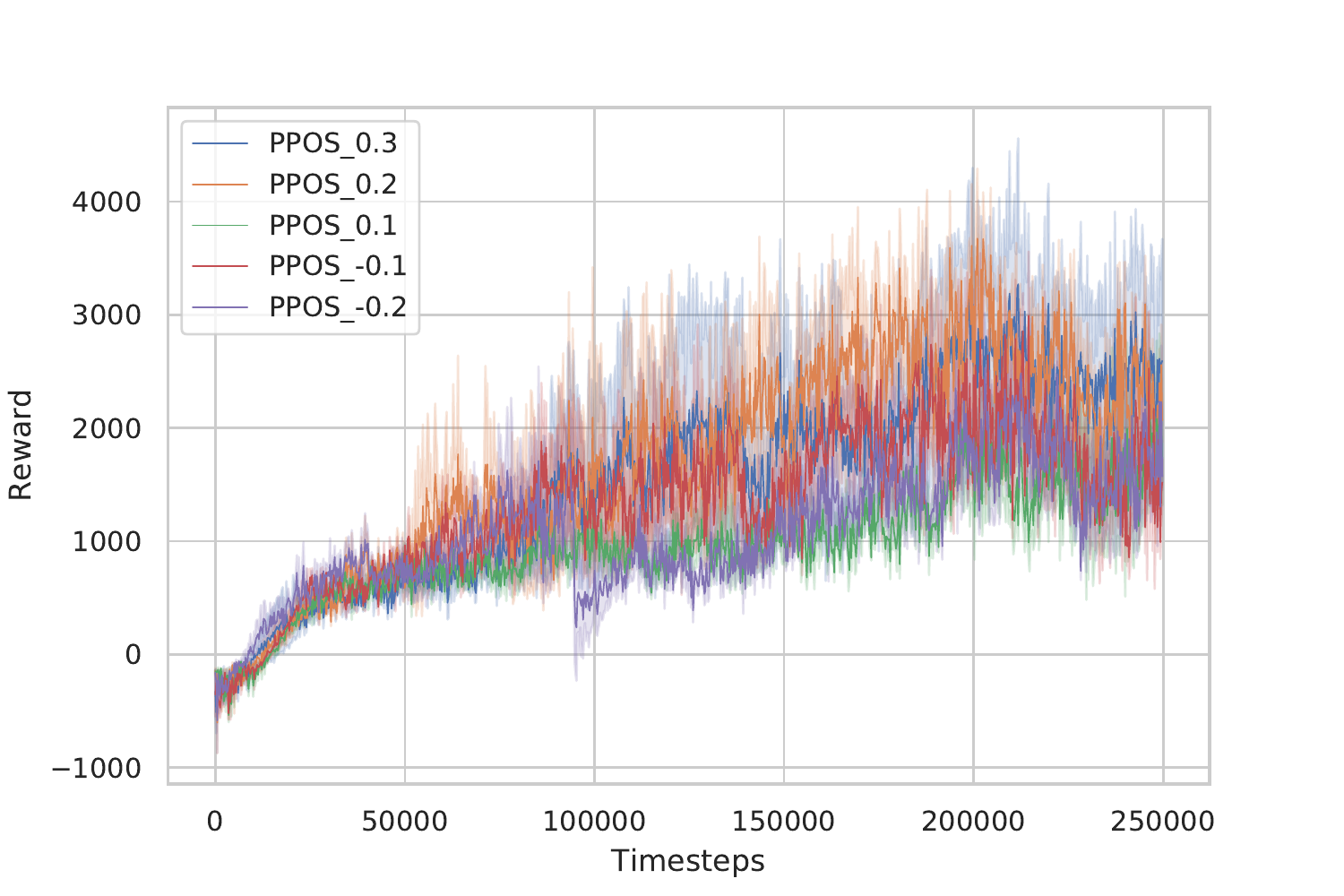}
}
 \caption{Episode rewards of the policy during the training process, averaged over ten random seeds on the Humanoid and the Ant tasks, in comparison with different $\alpha$ values of PPOS. The solid lines are mean values of ten trials while the shaded area depicts the 60$\%$ confidence interval. $|\mathcal{O}|$ represents the dimension of observation.}
\label{fig:alphaTask}
\end{figure}
\begin{table}[h]
\renewcommand\arraystretch{1.5}
\centering
\caption{Choices of $\alpha$ values for different control environments}
\label{tab:2}       
\begin{tabular}{p{5cm}|p{5cm}}
\hline\noalign{\smallskip}
Environment & Value   \\
\noalign{\smallskip}\hline\noalign{\smallskip}
Humanoid-v2 & 0.05 \\
Ant-v2 &  0.2\\
HalfCheetah-v2 & 0.3   \\
Swimmer-v2 & 0.3   \\
Reacher-v2 & 0.3 \\
\noalign{\smallskip}\hline
\end{tabular}
\end{table}

\begin{figure}
    \centering
\includegraphics[width=7cm]{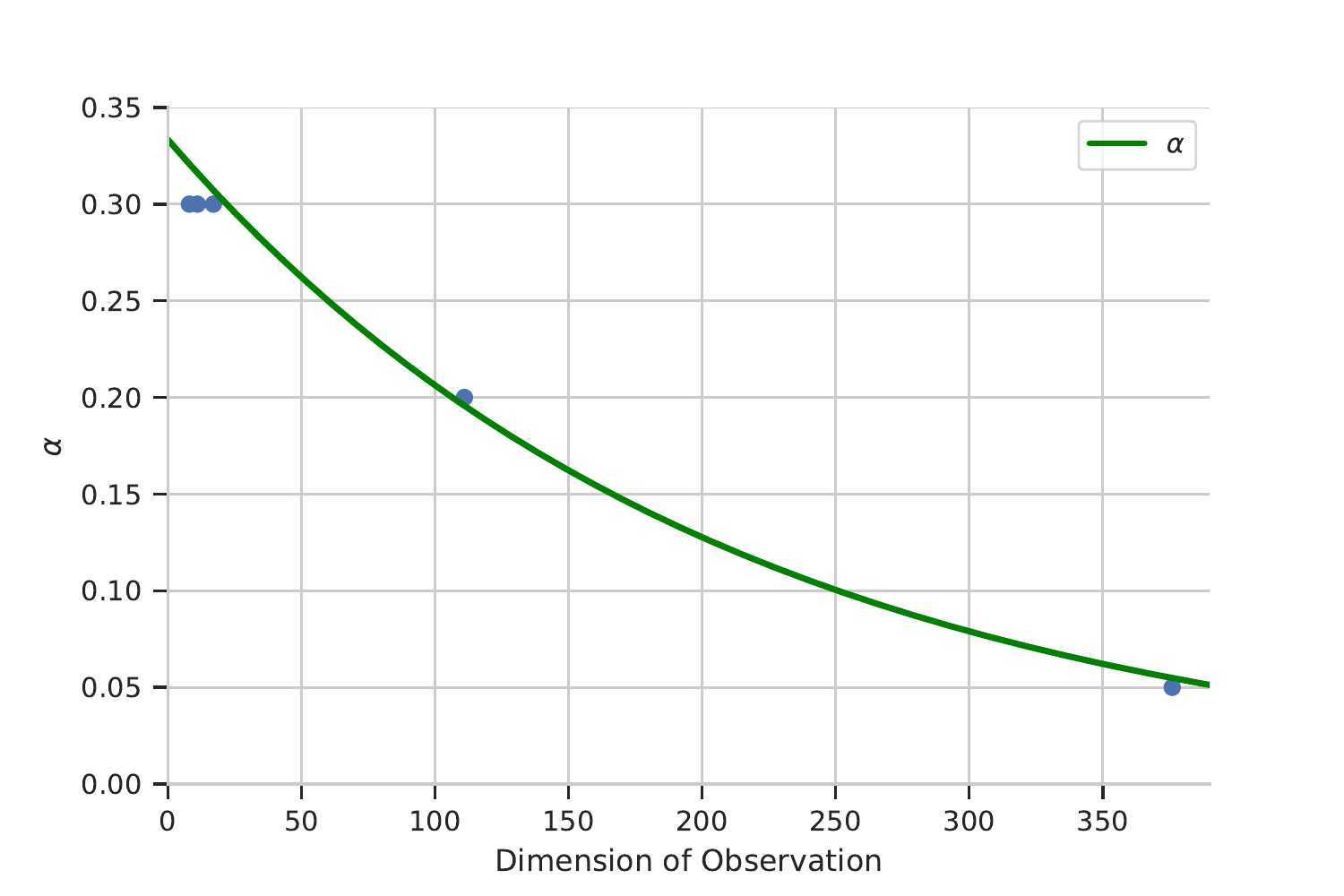}
    \caption{The proposed $\alpha$ values in relation to the dimension of observation, derived from the exponential regression of five MuJoCo tasks we conducted. Generally speaking, higher dimensions of observation compel us to choose a lower $\alpha$ value to achieve a better performance.}
    \label{fig:alphaDimension}
\end{figure}

Correct hyperparametric choices are crucial for policy search algorithms. For example, in PPORB the coefficient of the rollback operation $\alpha$ could vary from 0.3 to 0.05 for different tasks. However, the suggesting value was only provided by empirical methods for specific tasks in most of the former algorithms e.g. \cite{Ref9}, \cite{Ref11}, and \cite{Ref13}. With this in mind we evaluate our choices of  hyperparameters on varied tasks and provide a selection criteria of the hyperparameter value for different problem dimensionalities.\\
We conduct the experiment on Humanoid-v2 and Ant-v2 tasks, which have a higher dimensionality in their observation space, since in low-dimensional tasks the performance difference between PPO, PPORB and PPOS can be very subtle. As shown in Fig. \ref{fig:alphaTask}, in the Humanoid-v2 task the coefficient $\alpha=0.05$ achieves the best performance among different values, while $\alpha=0.2$ is the best choice in the Ant-v2 task. As discussed in section \ref{sec3.1}, the negative $\alpha$, which enables a smoother clipping function, is uncapable of improving the performance of the Ant-v2. A possible solution for this issue is decreasing the clipping range towards a lower $\epsilon$ value. For the other three tasks with relatively low dimensions, the default $\alpha=0.3$ can be regarded as the benchmark, which is presented in Table \ref{tab:2}.\\
Once we empirically found the optimal parameters for all five environments, we use an exponential regression function $|\mathcal{O}|=0.3333e^{-0.0048\alpha}$ to fit $\alpha$ value regarding the dimension of observation as shown in Fig. \ref{fig:alphaDimension}. In general, the proposed $\alpha$ is lower when the dimension of observation is higher. Our explanation for this is that a lower $\alpha$ value is effective in preventing unnecessarily large policy updates which are more likely to take place in high-dimensions. For low-dimensional tasks, default $\alpha=0.3$ can balance the performance and robustness. The exponential regression function has two key advantages compared with linear and polynomial regressions: Firstly, it converges to 0 as $\alpha$ grows with an elegant shape which meets the intuition that higher dimensions will result in lower $\alpha$ values. Secondly, the function is more efficient to express the same data than higher-order polynomial expressions.

\section{Conclusion}
Although PPO group algorithms have achieved state-of-the-art performance on policy searches on high-dimensional control tasks, they also present flaws when restricting the step of each policy update. The original clipping method of PPO can not prevent large policy updates, while the improved PPORB can not guarantee the convergence of this very same update. Additionally, as hyperparameters were proposed according to specific tasks, users are left without guidance to retune those parameters when redeploying the algorithm to their needs. Based on these observations, in this paper we proposed the PPOS algorithm with a functional clipping mechanism combined to the original PPO. Our work significantly improves the ability to restrict policy updates and, to maintain stability, and we demonstrated those through extensive results. We also provide a helpful guideline for tuning the only hyperparameter used by our method, and thus resulting in efficient redeployment on novel environments in the future.\\
While in this paper we focused on the hyperbolic tangent function and the tuning of the coefficient, in the future we will consider the clipping mechanism with different functions, as other functions have different properties which will affect the performance of the policy search. We will also study the relationship between the clipping range and the clipping function. The default $\epsilon=0.2$ is widely used in different variants of PPO, so there is considerable space for exploration of the clipping range. Another possible improvement in the future is the trust-region method, which is another mechanism to restrict the policy updates \cite{Ref9}. Finally, the implementations of deep reinforcement learning algorithms suffer from the repetitive tuning of hyperparameters \cite{Ref28}. Although we proposed a tuning method for one hyperparameter, there are still many hyperparameters, as seen in Table \ref{tab:3}, which may need to be explored in those frameworks.


\section{Appendix}
\begin{table}[h]
\renewcommand\arraystretch{1.5}
\centering
\caption{Hyperparameter of the control environments}
\label{tab:3}       
\begin{tabular}{p{5cm}|p{5cm}}
\hline\noalign{\smallskip}
Hyperparameter & Value   \\
\noalign{\smallskip}\hline\noalign{\smallskip}
$\epsilon$ & 0.2 \\
$\alpha$ in PPORB &  0.02 (Humanoid) , 0.3 (Other)\\
buffer size & 20000   \\
learning rate & 0.0003   \\
$\gamma$ & 0.99 \\
epoch  &  100  \\
step per epoch  & 2500 \\
collect per step & 10 \\
repeat per collect & 2 \\
batch size & 128 \\
hidden layer number & 1 \\
hidden neurons per layer & 128 \\
training number & 8 \\
test number & 100 \\
$\lambda$ (GAE) & 0.95 \\
policy & Gaussian \\
\noalign{\smallskip}\hline

\end{tabular}
\end{table}
\clearpage

\end{document}